\pdfoutput=1

\documentclass[11pt]{article}

\usepackage[final]{emnlp2021}

\usepackage{times}
\usepackage{latexsym}

\usepackage[T1]{fontenc}

\usepackage[utf8]{inputenc}

\usepackage{microtype}
\usepackage{tabularx}
\usepackage{booktabs}
\usepackage{multirow}
\usepackage{color}
\usepackage{amssymb}
\usepackage{bm}
\usepackage{graphicx}
\usepackage{algorithm}
\usepackage{amsmath}
\usepackage{algorithmicx}
\usepackage{CJK}
\usepackage{algpseudocode}
\usepackage{pifont}
\title{Constructing Emotion Consensus and Utilizing Unpaired Data for Empathetic Dialogue Generation}

\def\first{$^1$}
\def\second{$^2$}
\def\third{$^3$}
\def\comma{$^,$}
\def\star{$^*$}

\author{Lei Shen\first\comma\second 
~~~ Jinchao Zhang\third
~~~ Jiao Ou\first\comma\second
~~~ Xiaofang Zhao\first\star
~~~ Jie Zhou\third
\\
{\first {Institute of Computing Technology, Chinese Academy of Sciences, Beijing, China}}  \\
{\second {University of Chinese Academy of Sciences, Beijing, China}} \\
{\third {Pattern Recognition Center, WeChat AI, Tencent Inc, China}} \\
{\tt \small{\{shenlei17z, oujiao17b,zhaoxf\}@ict.ac.cn}}, {\tt \small{ \{dayerzhang,withtomzhou\}@tencent.com}}}

\begin{document}
\maketitle

\newcommand\blfootnote[1]{%
\begingroup 
\renewcommand\thefootnote{}\footnote{#1}%
\addtocounter{footnote}{-1}%
\endgroup
}
\blfootnote{Joint work with Pattern Recognition Center, WeChat AI, Tencent Inc, China.\star{Xiaofang Zhao is the corresponding author.}}

\begin{abstract}
Researches on dialogue empathy aim to endow an agent with the capacity of accurate understanding and proper responding for emotions. Existing models for empathetic dialogue generation focus on the emotion flow in one direction, that is, from the context to response. We argue that conducting an empathetic conversation is a bidirectional process, where empathy occurs when the emotions of two interlocutors could converge on the same point, i.e., reaching an \textit{emotion consensus}. Besides, we also find that the empathetic dialogue corpus is extremely limited, which further restricts the model performance. To address the above issues, we propose a dual-generative model, Dual-Emp, to simultaneously construct the emotion consensus and utilize some external unpaired data. Specifically, our model integrates a forward dialogue model, a backward dialogue model, and a discrete latent variable representing the emotion consensus into a unified architecture. Then, to alleviate the constraint of paired data, we extract unpaired emotional data from open-domain conversations and employ Dual-Emp to produce pseudo paired empathetic samples, which is more efficient and low-cost than the human annotation. Automatic and human evaluations demonstrate that our method outperforms competitive baselines in producing coherent and empathetic responses.
\end{abstract}

\section{Introduction}
\label{introduction}

\begin{figure}[htb]
\begin{center}
   \includegraphics[width=1.0\linewidth]{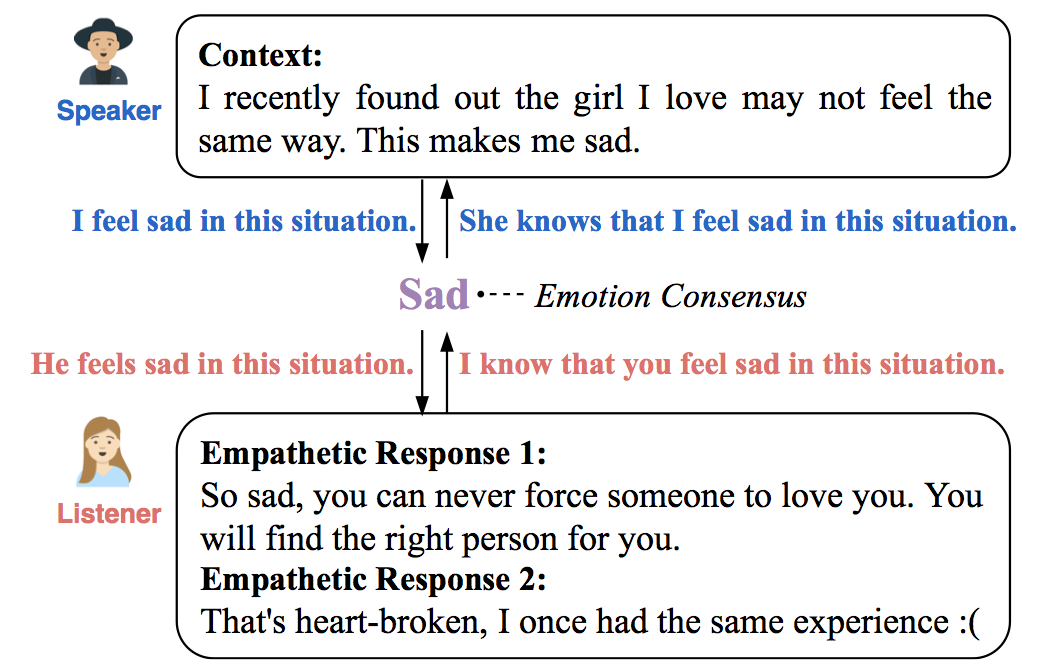}
\end{center}
   \caption{An example of conducting an empathetic conversation. Both responses show empathy to the speaker.}
\label{fig:intro}
\end{figure}

Empathy, a fundamental trait of humans, describes the ability to place oneself in another person's position and share his/her feelings or emotions. Besides, it has been considered to be one of the most valuable affective phenomena for improving human-machine interactions \cite{zech2005talking}. The studies of empathy in natural language processing mainly include detecting empathy in spoken language or text \cite{buechel2018modeling,sharma2020computational}, generating empathetic dialogue responses \cite{lin2019moel,majumder2020mime,sabour2021cem}, and constructing empathy lexicons \cite{sedoc2020learning} or datasets \cite{rashkin2019towards}.


The empathetic dialogue generation task has been regarded as a unidirectional process from the context to response, and is modeled as a multi-task learning that combines the emotion understanding and the emotion-enhanced response generation. Therefore, existing work \cite{rashkin2019towards,lin2019moel,li2020empathetic1,majumder2020mime} mainly focuses on improving the accuracy of emotion classification or enhancing response generation via integrating the detected emotion factor. 

Conducting an empathetic conversation is naturally a bidirectional process: the speaker conveys his/her emotion by describing a certain situation, then the listener receives that emotion and feeds his/her feeling back to the listener via a response. Then, the empathy is triggered when two interlocutors link similar experiences and their emotions could converge on the same point, i.e., reaching an \textit{emotion consensus}. 
Take the case in Figure \ref{fig:intro} as an example. The emotion consensus ``Sad'' works as an intersection that connects both the speaker and listener, and it is a high-level abstraction behind the content, i.e., both two responses convey their acknowledgment of the sad feeling even with different expressions. Therefore, a unidirectional model is not enough to model the relationship between the context and response. 
Besides, previous models for this task only utilize paired data with limited capacity in a benchmark dataset, \textsc{EmpatheticDialogues}. Rather than manually annotating a larger empathetic dataset, we find that in open-domain conversations, there is large-scale emotional data that can be used to improve the performance. Compared with recognizing whether a context-response pair is empathetic, obtaining either an emotional context or response (named as \textit{unpaired data} in this paper) can be easier with a well-trained classifier.

In this paper, we propose a \textbf{Dual}-Generative model for the \textbf{Emp}athetic dialogue generation task (Dual-Emp), which simultaneously constructs emotion consensus and utilizes unpaired data. Dual-Emp combines a forward dialogue model (generating a response based on its context) and a backward dialogue model (generating a context based on its responses) with a discrete latent variable. Specifically, the forward and backward encoders convert the context and response into vectors at the same time, and then a discrete latent variable is used to capture the high-level emotion consensus shared in each context-response pair. Moreover, the latent variable and an emotion-enhanced attention mechanism are integrated into both forward and backward decoders to better express proper empathy. 
To utilize unpaired emotional data, we firstly extract them from open-domain conversations with emotions. Then we can get pseudo pairs by feeding either emotional responses or contexts to the backward or forward model. A joint training process is introduced to promote the semantic coherence between contexts and responses. Furthermore, two types of optimization methods are applied to better train the entire model with paired and unpaired data. Experimental results on a benchmark dataset \textsc{EmpatheticDialogues} show that Dual-Emp significantly outperforms competitive baselines in generating meaningful and related responses while expressing an appropriate empathy.

Our main contributions can be summarized as: (1) We point out that the empathetic dialogue generation contains bidirectional processes, and highlight the importance of constructing emotion consensus. Besides, we propose a novel dual-generative model that couples a forward and a backward dialogue model with a discrete latent variable capturing the shared emotion consensus. (2) We utilize unpaired emotional data to break the constraint of paired empathetic data in the widely-used benchmark dataset \textsc{EmpatheticDialogues}. (3) Automatic and human evaluations show that our model outperforms competitive baselines in terms of fluency, coherence, and empathy.

\section{Related Work}
\label{relatedwork}

\noindent\textbf{Emotion-Controllable Response Generation.}
Infusing emotions into dialogue systems can make conversational agents more human-like and benefit the interactions between human and machine \cite{prendinger2005empathic}. Emotion-controllable response generation aims to generate emotional responses conditioning on a manually-provided label. Existing work \cite{zhou2018emotional,zhou2018mojitalk,colombo2019affect,song2019generating,shen2020cdl,zheng2021comae} focused on obtaining responses that are not only meaningful, but also in accordance with the desired emotion.

\noindent\textbf{Empathetic Response Generation.}
\citet{rashkin2019towards} considered a richer and evenly distributed set of emotions, and released a dataset \textsc{EmpatheticDialogues}. \citet{shin2020generating} formulated a reinforcement learning problem to maximize user’s sentimental feelings towards the generated responses. \citet{lin2019moel} presented an encoder-decoder model with each emotion having a dedicated decoder. \citet{majumder2020mime} introduced emotion grouping, emotion mimicry, and stochasticity to generate empathetic and various responses. \citet{li2020empathetic1} integrated knowledge to better understand dialogue contexts, and also designed an emotion-focused attention mechanism for emotional dependencies.

\noindent\textbf{Dual Learning in NLP.}
\citet{he2016dual} proposed Dual Learning (DL) for machine translation first, which considered the source to target language translation and target to source language translation as a dual task. After that, \citet{tang2017question} implemented a dual framework for the question-answering system. Both \citet{zhang2018reinforcing} and \citet{cui2019dal} used similar idea in dialogue generation task to produce coherent but not safe responses. \citet{shen2020cdl} applied DL for emotion-controllable response generation with three awards for emotions and semantics. Some researchers also exploited DL to relieve the need of paired data and make use of unpaired data in several areas, such are style transfer \cite{luo2019towards,luo2019dual}, semantic understanding \cite{tseng2020generative}, stylized response generation \cite{zheng2020stylized}, and machine translation \cite{zheng2019mirror}. 

The differences between our model and previous methods are: (1) To improve the empathy understanding, we introduce a backward model to represent the response and a discrete latent variable to capture the emotion consensus shared by contexts and responses. (2) Our forward and backward models are connected by a latent variable, and \textit{both of them can be updated at each iteration, while traditional DL can only fix one to update another}.

\section{Proposed Method}
\label{method}

For empathetic dialogue generation, a dialogue consists of utterances from a speaker and a listener. Given context $\textbf{c} = \{S_1, L_1, S_2, L_2., ..., S_t\}$, where $S_i={\{w_j^i\}_{j=1}^{|S_i|}}$ denotes speaker and $L_i={\{w_j^i\}_{j=1}^{|L_i|}}$ denotes listener, the goal is to track the speaker's emotion state from $\textbf{c}$, and generate a response $\textbf{y}=L_t$ that is meaningful and empathetic.

\subsection{Overview}
\begin{figure}[htb]
\begin{center}
   \includegraphics[width=0.9\linewidth]{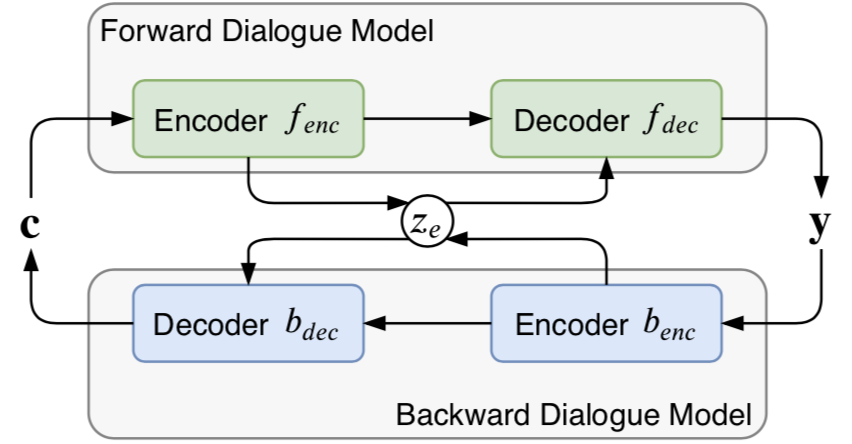}
\end{center}
   \caption{The architecture of Dual-Emp. It couples forward and backward dialogue models with a discrete latent variable $z_e$ denoting emotion consensus.}
\label{fig:overview}
\end{figure}

The architecture of Dual-Emp is shown in Figure \ref{fig:overview}. Dual-Emp has five modules, the forward encoder $f_{enc}$, forward decoder $f_{dec}$, backward encoder $b_{enc}$, backward decoder $b_{dec}$, and $z_e$ indicating a discrete latent variable. $z_e$ can be inferred from both $\textbf{c}$ and $\textbf{y}$ and is used to capture emotion consensus shared in each $\langle\textbf{c}$, $\textbf{y}\rangle$ pair. Because of the existence of $z_e$, other modules are correlated and can better model both the semantic relation and the emotion connection between $\textbf{c}$ and $\textbf{y}$.

\subsection{Model Architecture}
Since the backward dialogue model has the same architecture as the forward one, we specify the components of forward dialogue model below and omit those of backward model for space limitation.

\noindent\textbf{Encoder.}
Following the work of \citet{lin2019moel}, we firstly concatenate utterances in $\textbf{c}$ into a long sequence with length $n$ and add a special token $\mathrm{CTX}$ to the beginning of $\textbf{c}$ inspired by BERT \cite{devlin2019bert}. Then, each token $w$ in $\textbf{c}$ is calculated as the sum of three embeddings:
\begin{equation}
    \textbf{E}_c(w) = \textbf{ E}_w(w) + \textbf{ E}_p(w) + \textbf{ E}_r(w),
\end{equation}
where $\textbf{E}_w(\cdot)$, $\textbf{E}_p(\cdot)$, and $\textbf{E}_r(\cdot) \in \mathbb{R}^{|V| \times d_{emb}}$ represent word embedding space, positional embedding space and role embedding space\footnote{The roles in $\textbf{c}$ is an alternating set of ``speaker'' and ``listener'', while in $\textbf{y}$, the role is ``listener'' only.}, respectively. Finally, a transformer encoder \cite{vaswani2017attention}  $f_{enc}$ is applied to get the context representation:
\begin{equation}
    \textbf{H} = f_{enc}(\textbf{E}_c([\mathrm{CTX};\textbf{c}])),
\end{equation}
where ``$;$'' represents the concatenation operation, and $\textbf{H}\in \mathbb{R}^{(n+1) \times d_{mod}}$. The contextualized encoding of $\mathrm{CTX}$, i.e., $\textbf{H}_0 \in \mathbb{R}^{d_{mod}}$, is used as the final representation of the entire context.

\noindent\textbf{Emotion Consensus Construction.}
A $K$-way categorical latent variable $z_e \in [1, K]$ \cite{bao2019plato} is used to capture the emotion consensus shared by $\textbf{c}$ and $\textbf{y}$.
Inspired by \citet{zhao2019rethinking}, we define the prior distribution where we sample $z_e$ from to be uniform\footnote{Since emotion labels in \textsc{EmpatheticDialogues} are evenly-distributed, we set the prior distribution to be uniform.}, i.e., $p(z_{e})= 1/K$. Correspondingly, the approximate posterior distribution is defined as follows:
\begin{equation}
    q(z_e|\textbf{c}) = \mathrm{softmax}(\mathrm{FFN}(\textbf{H}_0)) \in \mathbb{R}^K,
\end{equation}
where $\mathrm{FFN}(\cdot)$ represents a feedforward network. This part can be considered as the emotion understanding on $c$. Here $z_e$ has its own embedding space $\textbf{E}_z \in \mathbb{R}^{K \times d_{mod}}$ to convert it into a vector, i.e., $\textbf{E}_{[z]} =\textbf{E}_z(z_e) \in \mathbb{R}^{d_{mod}}$. To supervise the emotion expression in $\textbf{E}_{[z]}$, we train a classifier using the cross-entropy loss between $\textbf{E}_{[z]}$ and ground-truth emotion label $e^{*}$:
\begin{align}
     &p_e = \mathrm{softmax}(\textbf{W}_e\textbf{E}_{[z]}), \\ 
     &\mathcal{L}_{emo} = -e^{*}\log p_e,
\label{45}
\end{align}
where $\textbf{W}_e$ is a trainable weight matrix.

\begin{figure*}[htb]
\begin{center}
   \includegraphics[width=1.0\linewidth]{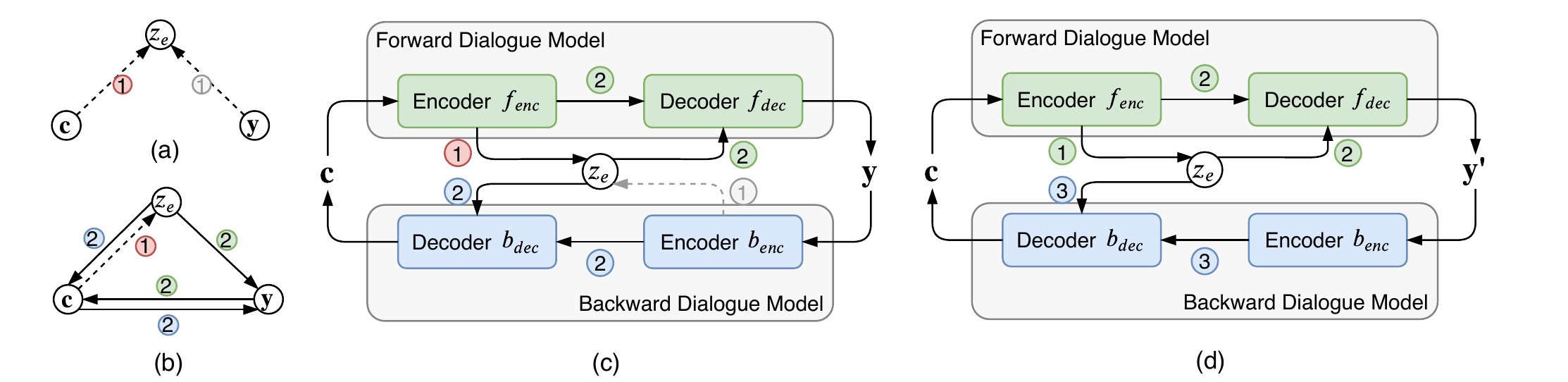}
\end{center}
   \caption{Illustration of the training process. (a) shows the inference of $z_e$. Both $\textbf{c}$ and $\textbf{y}$ are used to infer $z_e$ that represents the emotion consensus shared by $\textbf{c}$ and $\textbf{y}$. (b) shows the graphical model of (c), and (c) depicts procedures to compute $\mathcal{L}_1$ (Eq. \ref{l_cy})/$\mathcal{L}_2$ (Eq. \ref{l_cy2}). (d) represents procedures to compute $\mathcal{L}_{3}$ (Eq. \ref{l_c}).} 
\label{fig:optimization}
\end{figure*}

\noindent\textbf{Decoder.}
Existing work \cite{lin2020variational,li2020empathetic1} mainly integrates the obtained emotion factor to either the first decoding position or all steps. To focus on emotion consensus dynamically, we apply an emotion-enhanced attention mechanism in the cross-attention layer of transformer decoder. We firstly concatenate $\textbf{E}_{[z]}$ with token embeddings of the decoder input $\{y_i\}_{i=1}^{t-1}$ to get representations $\textbf{Y} = \{\textbf{y}_i\}_{i=0}^{t-1}$ with $\textbf{y}_0 = \textbf{E}_{[z]}$. Then we feed $\textbf{Y}$ into decoder $f_{dec}$. 

Our decoder has similar structure to the transformer decoder. The input $\textbf{Y}$ is converted to $\textbf{D}$ by the self-attention layer. As $\textbf{H}$ and $\textbf{E}_{[z]}$ serve different purposes, we design a cross-attention layer with two separate key-value matrices, and the encoder-decoder vectors are computed as follows:
\begin{align}
    \textbf{C}_H &= \mathrm{MultiHead}(\textbf{D}, \textbf{H}, \textbf{H}), \\
    \textbf{C}_Z &= \mathrm{MultiHead}(\textbf{D}, \textbf{E}_{[z]}, \textbf{E}_{[z]}),
\end{align}
where $\mathrm{MultiHead}(\textbf{Q}, \textbf{K}, \textbf{V})$ is a multi-head attention function taking a query matrix $\textbf{Q}$, a key matrix $\textbf{K}$, and a value matrix $\textbf{V}$ as inputs. The fully connected feedforward layer is defined as:
\begin{equation}
    \hat{\bf Y} = \mathrm{FFN}([\textbf{C}_H; \textbf{C}_Z]),
\end{equation}
where $\hat{\bf Y}=\{\hat{\bf y}_i\}_{i=1}^t$. Finally, the decoding distribution over the vocabulary of the next token is computed as:
\begin{equation}
    p(y_t|y_{<t}, \textbf{c}, z_e) = \mathrm{softmax}(\textbf{W}_o \hat{\bf y}_t),
\end{equation}
where $\textbf{W}_o$ is a trainable weight matrix.

\subsection{Training and Inference}
We firstly describe how Dual-Emp can be trained with the paired data $\langle \textbf{c}, \textbf{y} \rangle$, and also the unpaired data \textbf{c} or \textbf{y}. Then a combined objective is derived to optimize Dual-Emp using the paired and unpaired data at the same time.

\noindent\textbf{Training with Paired Data.}
Given $\langle\textbf{c}, \textbf{y}\rangle$, we aim to maximize the log-likelihood of a joint probability $p(\textbf{c}, \textbf{y})$:
\begin{equation}
    \log p(\textbf{c}, \textbf{y}) = \log \sum\nolimits_{z_e}p(\textbf{c}, \textbf{y}, z_e).
\end{equation}
Following the derivations from \citet{zhao2018unsupervised}, \citet{zhao2019rethinking}, \citet{tseng2020generative}, and the variational inference \cite{kingma2013auto}, an objective based on the evidence lower bound can be derived as:
\begin{equation}
\begin{aligned}
    \mathcal{L}_1 = & \mathbb{E}_{q(z_e|\textbf{c})} \log p(\textbf{y}|z_e, \textbf{c}) + \mathbb{E}_{q(z_e|\textbf{c})} \log p(\textbf{c}|z_e, \textbf{y}) \\   &-D_\mathrm{KL}[q(z_e|\textbf{c})||p(z_e)],
\end{aligned}\label{l_cy}
\end{equation}
where the first term denotes the forward dialogue model, $q(z_e|\textbf{c})$ is the approximate posterior distribution of $z_e$, and is computed by the forward encoding process (red \ding{172} in Figure \ref{fig:optimization}(c)). $p(\textbf{y}|z_e, \textbf{c})$ is the forward decoding process (green \ding{173} in Figure \ref{fig:optimization}(c)); the second term denotes the reconstruction of $\textbf{c}$, and $p(\textbf{c}|z_e, \textbf{y})$ is the backward decoding process (blue \ding{173} in Figure \ref{fig:optimization}(c)); the third term is a Kullback-Leibler (KL) divergence between two distributions.

Analogously, the posterior distribution of $z_e$ can be approximated by $q(z_e|\textbf{y})$, and the objective can be converted as follows:
\begin{equation}
\begin{aligned}
    \mathcal{L}_2 = & \mathbb{E}_{q(z_e|\textbf{y})} \log p(\textbf{c}|z_e, \textbf{y}) + \mathbb{E}_{q(z_e|\textbf{y})} \log p(\textbf{y}|z_e, \textbf{c}) \\   &-D_\mathrm{KL}[q(z_e|\textbf{y})||p(z_e)],
\end{aligned}\label{l_cy2}
\end{equation}
where terms have similar meanings to those in Eq. \ref{l_cy}, and we only need to interchange ``forward'' and ``backward''. Besides, the forward encoding process (red \ding{172} in Figure \ref{fig:optimization}(c)) is replaced with the backward encoding process (gray \ding{172} in Figure \ref{fig:optimization}). Detailed derivations can be found in Appendix. Therefore, the final loss function for the paired data is:
\begin{equation}
    \mathcal{L}_{cy} = \mathcal{L}_1 + \mathcal{L}_2 + \alpha\mathcal{L}_{emo}.
\end{equation}
where $\alpha$ is a hyper-parameter.

\noindent\textbf{Training with Unpaired Data.}
Given unpaired data $\textbf{c}$ ($\textbf{c}$ is an emotional context), we need to maximize the log-likelihood of a marginal probability $p(\textbf{c})$:
\begin{equation}
    \log p(\textbf{c}) = \log \int_\textbf{y}\sum\nolimits_{z_e} p(\textbf{c}, \textbf{y}, z_e).
\end{equation}
Then, we can get the evidence lower bound for the marginal probability:
\begin{equation}
\begin{aligned}
    \mathcal{L}_3 = &\mathbb{E}_{q(\textbf{y}|z_e, \textbf{c})}\mathbb{E}_{q(z_e|\textbf{c})}\log p(\textbf{c}|z_e, \textbf{y}) \\
    & -D_\mathrm{KL}[q(z_e|\textbf{c})||p(z_e)],
\end{aligned}
\label{l_c}
\end{equation}
where the first term is the reconstruction of $\textbf{c}$, $q(z_e|\textbf{c})$ is computed by the forward encoding process (\ding{172} in Figure \ref{fig:optimization}(d)), $q(\textbf{y}|z_e, \textbf{c})$ is the forward generation process (\ding{173} in Figure \ref{fig:optimization}(d)), and $p(\textbf{c}|z_e$, $\textbf{y})$ is the backward decoding process (\ding{174} in Figure \ref{fig:optimization}(d)); the second term is a KL divergence. 

The forward generation process $q(\textbf{y}|z_e, \textbf{c})$ is similar to the back-translation in machine translation \cite{zhang2018joint}, and we use $f_{dec}$ to generate pseudo $\textbf{y'}$ given $\textbf{c}$ and $z_e$. Since the ground-truth $\textbf{y}$ is unobserved here, we apply reinforcement learning and policy gradient method \cite{williams1992simple} for training. The reward is designed as the probability of the model to reconstruct $\textbf{c}$ based on the generated $\hat{\textbf{y}}$ and $z_e$:
\begin{equation}
    r = p(\textbf{c}|\hat{\textbf{y}}, z_e).
\end{equation}

Similarly, we can get an objective when utilizing unpaired $\textbf{y}$ (the emotional response):
\begin{equation}
\begin{aligned}
    \mathcal{L}_4 = &\mathbb{E}_{q(\textbf{c}|z_e, \textbf{y})}\mathbb{E}_{q(z_e|\textbf{y})}\log p(\textbf{y}|z_e, \textbf{c}) \\
    & -D_\mathrm{KL}[q(z_e|\textbf{y})||p(z_e)],
\end{aligned}
\end{equation}
where the first term is the reconstruction of $\textbf{y}$, and the process is symmetrical to that of $\mathcal{L}_3$. Detailed derivations can be found in Appendix. The final loss functions for unpaired data $\textbf{c}$ and $\textbf{y}$ are:
\begin{align}
     &\mathcal{L}_c = \mathcal{L}_3 + \beta\mathcal{L}_{emo}, \\
     &\mathcal{L}_y = \mathcal{L}_4 + \gamma\mathcal{L}_{emo},
\end{align}
where $\beta$ and $\gamma$ are two hyper-parameters.

\noindent\textbf{Total Training Loss.}
During training, the paired empathetic data in \textsc{EmpatheticDialogues} and the unpaired emotional data from open-domain conversations are used simultaneously. Then, the total loss can be summarized as:
\begin{equation}
    \mathcal{L} = \mathcal{L}_{cy} + \mathcal{L}_c + \mathcal{L}_y.
\end{equation}

\noindent\textbf{Inference.}
During inference, given the input $\textbf{c}$, only the forward dialogue model is applied. We use $f_{enc}$ to encode $\textbf{c}$ and infer $z_e$, then employ $f_{dec}$ to generate $\hat{\textbf{y}}$ based on $\textbf{c}$ and $z_e$.

\section{Experiments}
In this section, we conduct experiments to evaluate our proposed method. We firstly introduce some empirical settings. Then we illustrate our results on both automatic and human evaluations. Finally, we show some cases generated by different models and do further analyses over our method.

\subsection{Dataset}
We conduct our experiments on the \textsc{EmpatheticDialogues} \cite{rashkin2019towards} dataset that consists of 24,850 conversations between two interlocutors. Each conversation in the dataset contains \textbf{one} emotion label, a situation where the speaker feels the exact emotion, and utterances about the speaker's descriptions of the situation or the listener's empathetic replies. There are 32 evenly-distributed emotion labels in the dataset. We apply the data provided by the original paper with the split ratio of 8:1:1 for training/validation/test set, and use the script released by \citet{lin2019moel} to preprocess the data. Emotion labels are given as supervised signals in the training process, while during inference, they are predicted to evaluate the accuracy of emotion understanding.

\subsection{Implementation Details}
We optimize the models using Adam \cite{kingma2014adam} with a mini-batch size of 16. The learning rate is initialized to 1e-4 and we vary the learning rate following \citet{vaswani2017attention}. 
Similar to \citet{lin2019moel}, \citet{li2020empathetic1}, and \citet{majumder2020mime}, we use pre-trained GloVe vectors \cite{pennington2014glove} to initialize the word embeddings. Besides, all common hyper-parameters are set the same as previous work, e.g., the hidden size $d_{mod}$ and embedding size $d_{emb}$ are set to 300. In order to alleviate the degeneration problem of variational framework, we apply KL annealing \cite{bowman2015generating} that is the same as in \citet{zhou2018emotional}. During inference, we use greedy decoding strategy and the maximum decoding step is set to 30. $K$ equals to 32. $\alpha$, $\beta$, and $\gamma$ are simply set to 1. The running epoch is set to 30 with early stopping.

To get unpaired emotional data, we utilize two large-scale datasets of open-domain conversations provided by \citet{zhong2020care}, namely Reddit and Twitter. 
Following \citet{zhou2018emotional} and \citet{shen2020cdl}, an emotion classifier is applied to obtain the ground-truth emotion label for each context and response. Here, we use the pre-trained classifier provided by \citet{rashkin2019towards} to predict labels among 32 emotions. The classifier could return one emotion label with the highest probability. We firstly keep contexts or responses with the probability larger than a threshold $s=0.60$. Then, we remove contexts or responses with length smaller than 3. Finally, 155,059 contexts and 149,672 responses are obtained as unpaired emotional data.

We use Pytorch\footnote{\url{https://pytorch.org/}} to implement the codes, and our model is trained on a Titan Xp GPU with an average running time of 2 days.

\subsection{Baselines}
We compare our approach with five representative
baselines: 
(1) \textbf{Multi-TRS} \cite{rashkin2019towards}: A transformer-based model trained with emotion classification loss in addition to MLE loss, and the emotion label is classified from the encoder output; (2) \textbf{MoEL} \cite{lin2019moel}: An extension to Multi-TRS, which softly combines the output states of the appropriate decoders and generates an empathetic response. Each decoder is optimized to focus on a specific emotion; (3) \textbf{EmpDG} \cite{li2020empathetic1}: A model that exploits coarse- and fine-grained emotions and introduces an interactive adversarial learning framework to use user feedbacks; (4) \textbf{DualVAE} \cite{tran2018dual}: A model with two decoders: one is for CVAE, and the other is for response auto-encoding; (5) \textbf{MIME} \cite{majumder2020mime}: A model that integrates emotion grouping, emotion mimicry, and stochasticity strategies to generate varied responses. MIME is also the state-of-the-art model for empathetic response generation. To make fair comparisons, we do not apply methods based on pre-trained models here, as both Dual-Emp and the above mentioned ones are not based on pre-trained models. Note that model (1) to (5) can only utilize the paired data.

Additionally, we also design following models for ablation study: (6) \textbf{Sing-Emp-Paired}: A variation of Dual-Emp with only the forward model and paired empathetic data; (7) \textbf{Dual-Emp-Paired}: Dual-Emp with only paired empathetic data.

\subsection{Evaluation Measures}

\begin{table*}[!htb]
    \small
    \centering
    \begin{tabular}{l|cccccccc|ccc}
    \hline
        \multirow{2}*{\textbf{Method}} & \multicolumn{8}{c|}{\textbf{Automatic Evaluation}} & \multicolumn{3}{c}{\textbf{Human Evaluation}}  \\ \cline{2-12}
        & B & Avg & Gre & Ext & PPL & D1 (\%) & D2 (\%) & EA (\%) & Emp & Rel & Flu \\ 
    \hline
        Multi-TRS & 2.56 & 0.938 & 0.786 & 0.541 & 33.82 & 0.68 & 2.62 & 35.17 & 3.08 & 3.21  & 3.14\\
        MoEL & 2.80 & 0.945 & 0.793 & 0.537 & 37.81 & 0.56 & 2.70 & 35.38 & 3.35 & 3.65 & 3.26\\
        Emp-DG & 2.79 & 0.935 & 0.788 & 0.532 & 34.31 & 0.47 & 2.10 & 34.35 & 3.27 & 3.54 & 3.38\\
        DualVAE & 2.76 & 0.941 & 0.791 & 0.540 & 33.46 & 0.77 & 3.21 & 35.36 & 3.36 & 3.62 & 3.45\\
        MIME & 2.82 & 0.946 & 0.794 & 0.536 & 37.53 & 0.51 & 2.68 & 34.88 & 3.40 & 3.79 & 3.41 \\
    \hline
        Sing-Emp-Paired & 2.77 & 0.944 & 0.790 & 0.533 & 32.71 & 0.75 & 2.91 & 28.75 & 3.57 & 3.77 & 3.49 \\
        Dual-Emp-Paired & 2.86 & 0.950 & 0.792 & 0.542 & 32.56 & 0.80 & 3.09 & 36.82 & 3.62 & 3.86 & 3.57 \\
        Dual-Emp & \textbf{2.91} & \textbf{0.957} & \textbf{0.796} & \textbf{0.545} & \textbf{31.01} & \textbf{1.08} & \textbf{3.23} & \textbf{37.53} & \textbf{3.82}& \textbf{4.08} & \textbf{3.62}\\
    \hline
    \end{tabular}
    \caption{Automatic and human evaluation results. The metrics BLEU, Average, Greedy, Extrema, Dist-1, Dist-2, Emotion accuracy, Empathy, Relevance, and Fluency are abbreviated as B, Avg, Gre, Ext, D1, D2, EA, Emp, Rel, and Flu, respectively. Results show that Dual-Emp achieves the best performance on all metrics, especially a large improvement in Dist-1/2, Emotion accuracy, and Empathy.}
    \label{tab:autohuman}
\end{table*}

\begin{table}[h]
\small
\centering
\begin{tabular}{lcccc}
  \hline
  \textbf{Dual-Emp} vs. & Win & Loss & Tie & Kappa \\
  \hline
  Multi-TRS & 43\% & 27\% & 30\% & 0.563 \\
  MoEL & 37\% & 32\% & 31\% & 0.548\\
  Emp-DG & 40\% & 28\% & 32\% & 0.506\\
  DualVAE & 39\% & 30\% & 31\% & 0.527\\
  MIME & 36\% & 32\% & 32\% & 0.569\\
  \hline
\end{tabular}
\caption{Results of human A/B test. Pairwise comparisons show that responses from Dual-Emp are more preferred by humans than those from baselines.}
\label{tab:ab}
\end{table}

\noindent\textbf{Automatic Metrics.}
For automatic evaluation, we use followings metrics: (1) BLEU \cite{papineni2002bleu}; (2) Embedding-based scores (Average, Greedy, and Extrema)\footnote{We employ a popular NLG evaluation project available at \url{https://github.com/Maluuba/nlg-eval}.} \cite{liu2016not,serban2017hierarchical}; (3) Perplexity (PPL) \cite{vinyals2015neural}; (4) Dist-1/2 \cite{li2016diversity}; (5) Emotion accuracy (the agreement between the ground-truth emotion labels and the predicted ones from Eq. \ref{45}). Emotion accuracy can be used to measure the ability of emotion understanding.

\noindent\textbf{Human Evaluation.}
Firstly, we randomly sample 100 contexts and their corresponding responses from our model as well as the baselines. Next, we send pairs of the context and generated response from different models to three professional annotators without order. Annotators are asked to evaluate each pair independently based on three distinct metrics: Empathy, Relevance, and Fluency \cite{rashkin2019towards,lin2019moel,majumder2020mime}. Empathy measures the degree of emotional understanding of context shown by the response; Relevance evaluates whether the generated responses are relevant on topic with the context; Fluency assesses the grammatical correctness and readability of the generated responses. Each metric is rated on five-scale with ``5'' represents the best performance. 

\noindent\textbf{Human A/B Test.}
In this part, we try to directly compare Dual-Emp with other baselines. We randomly sample 100 dialogues each for Dual-Emp vs. $\{$Multi-TRS, MoEL, EmpDG, DualVAE, MIME$\}$. Three annotators are given generated responses from either Dual-Emp or $\{$Multi-TRS, MoEL, EmpDG, DualVAE, MIME$\}$ in random order, and are asked to choose the better response. They can either choose one of the responses or select ``Tie'' when the provided options are either both good or both bad. The result of each sample is determined by majority voting. Finally, we calculate the percentage of samples where the first or second model generates the better response and where these two models perform similarly.

\subsection{Experimental Results}
\noindent\textbf{Automatic Evaluation Results.}
The automatic evaluation results are shown in the left part of Table \ref{tab:autohuman}. The top part is the results of all baseline models, and we can see that Dual-Emp outperforms other methods on all metrics (t-test, $p$-value $<$ 0.05). The improvements of Dual-Emp on \textit{PPL}, \textit{Dist-1/2}, and \textit{Emotion accuracy} are significant, indicating that it can improve emotion understanding, and also enhance content fluency and diversity simultaneously. MoEL, Emp-DG, and MIME have similar performance, as they try to either improve the emotion understanding or intensify the emotion-based response generation. 

The bottom part of Table \ref{tab:autohuman} shows the results of our ablation study. Comparisons between Sing-Emp-Paired and Dual-Emp-Paired show the effectiveness of capturing emotion consensus with the assistance of both backward model and discrete latent variable. Especially, the noticeable improvement of \textit{Emotion accuracy} indicates the discrete latent variable used for emotion prediction can help better model the emotion consensus by taking contexts and responses into consideration. In addition, we can find that with the support of unpaired emotional data, Dual-Emp achieves better results than Dual-Emp-Paired.

\noindent\textbf{Human Evaluation Results.}
Human evaluation in Table \ref{tab:autohuman} illustrates that Dual-Emp obtains the best performance (t-test, $p$-value $<$ 0.05) on all scores. This suggests that our bidirectional model with latent variable helps construct emotion consensus shared by contexts and responses, thus improving the topic consistency and evoking more empathetic expressions. Besides, as more unpaired emotional data is utilized, Dual-Emp can achieve better \textit{Fluency}. Additionally, we carry out pairwise comparisons to directly compare the response quality in Table \ref{tab:ab}. The results confirm that responses from Dual-Emp are more preferred by humans. Agreements to measure the consistency among three annotators are calculated with Fleiss' kappa \cite{fleiss1973equivalence}, and the kappa values indicate ``moderate agreement'' in our cases.

\subsection{Case Study}
Table \ref{tab:case} shows two examples generated by Dual-Emp and other baselines. In the first case, Dual-Emp generates the most context-consistent response with a proper “apprehensive” emotion by replying with words “scary” and “what happened", whereas baselines fail to understand the negative emotion or express inappropriate contents. In the second case, Dual-Emp generates a coherent and informative response, which corresponds to a subtle emotion change of the context from ``lost a job'' to ``hoping he can find a full time job soon''. The response is not only emotion-related, but also contains the correct personal pronoun ``he'' and keyword ``job''.

\begin{table*}[!htb]
    \small
    \centering
    \begin{tabular}{c|l|l}
    \hline
        \multirow{7}*{1} & \textbf{Emotion} & Apprehensive \\ \cline{2-3}
        & \textbf{Context} & I went skydiving last summer with my partner. It was so scary !\\
        & Ground-truth Response & Wow, that is scary. Were you nervous? \\ \cline{2-3}
        & MoEL & I am so sorry to hear that. I am glad you are okay!\\
        & Emp-DG & Did you go check it out? \\
        & MIME & Oh no! I am sorry to hear that. I hope you get it!\\
        & \textbf{Dual-Emp} & That is \textcolor[rgb]{0,0,0.7}{scary}! \textcolor[rgb]{0,0,0.7}{What happened} to you? \\
    \hline
        \multirow{7}*{2} & \textbf{Emotion} & Hopeful \\ \cline{2-3}
        & \textbf{Context} & My husband lost a job, but I am hoping he can find a full time job soon. \\
        & Ground-truth Response & He will, I have faith. \\ \cline{2-3}
        & MoEL & I am sorry to hear that. I am sure he will be fine. \\
        & Emp-DG & I am sorry to hear that. I hope you have a good time! \\
        & MIME & That is a good thing to do. I hope you have a good job! \\
        & \textbf{Dual-Emp} & That is unfortunate. \textcolor[rgb]{0,0,0.7}{I hope he gets a new job}. \\
    \hline
    \end{tabular}
    \caption{Generated responses from MoEL, Emp-DG, MIME, and Dual-Emp. In Case 1, Dual-Emp generates the most context-consistent response with a proper ``apprehensive" emotion by replying with words ``scary" and ``what happened". In Case 2, Dual-Emp captures a subtle emotion change of the context from ``lost a job'' to ``hoping he can find a full time job soon''. Besides, it contains the correct personal pronoun ``he'' and keyword ``job''.}
    \label{tab:case}
\end{table*}

\begin{figure*}[htb]
\begin{center}
   \includegraphics[width=1.0\linewidth]{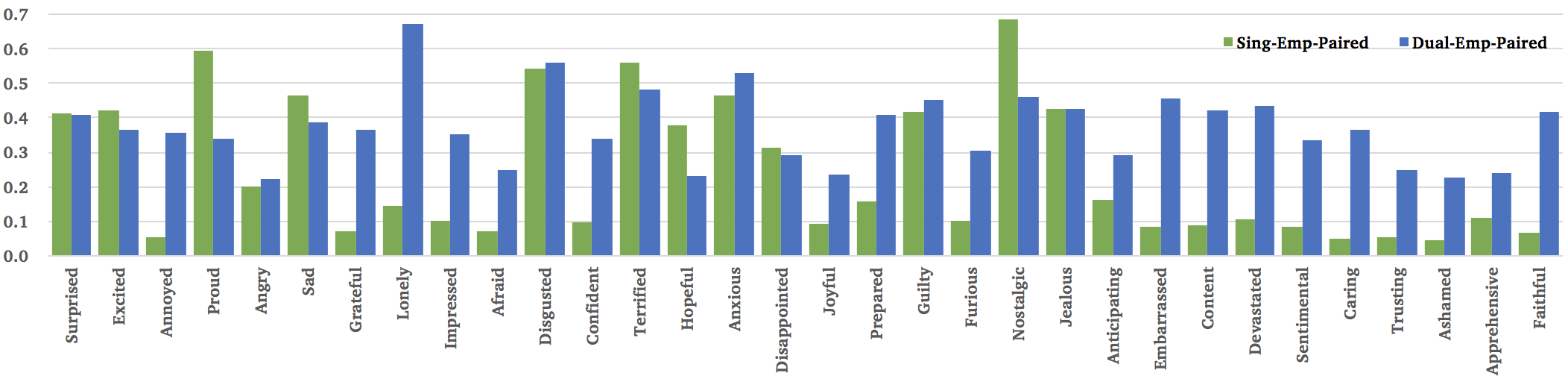}
\end{center}
   \caption{Emotion accuracy over 32 emotions of Sing-Emp-Paired and Dual-Emp-Paired. The accuracy of Sing-Emp-Paired is unbalanced among all emotions, while Dual-Emp-Paired can not only improve the overall accuracy, but also exhibit a relatively even performance.}
\label{fig:ana1}
\end{figure*}

\begin{table*}[!htb]
    \small
    \centering
    \begin{tabular}{l|cc|cccccccc}
    \hline
        $s$ & $\#$context & $\#$response & B & Avg & Gre & Ext & PPL & D1 (\%) & D2 (\%) & EA (\%) \\ 
    \hline
        0.50 & 324,243 & 314,070 & 2.25 & 0.937 & 0.791 & 0.549 & 31.22 & 0.90 & 2.90 & 35.62 \\ 
        0.55 & 224,324 & 216,839 & 2.69 & 0.933 & 0.784 & 0.537 & 32.63 & 1.87 & 4.32 & 36.06 \\
        0.60 & 155,059 & 149,672 & 2.91 & 0.957 & 0.795 & 0.545 & 31.01 & 1.08 & 3.23 & 37.53
        \\
        0.65 & 107,189 & 103,192 & 2.25 & 0.934 & 0.783 & 0.539 & 32.61 & 1.70 & 4.80 & 35.66 \\
        0.70 & 73,132 & 70,200 & 2.60 & 0.938 & 0.791 & 0.540 & 31.66 & 0.70 & 2.40 & 37.51 \\
    \hline
    \end{tabular}
    \caption{Automatic evaluation results based on the number of unpaired data with different $s$ values. Results show that more unpaired data does not lead to better results as some labels are not adequate with a low confidence.}
    \label{tab:s}
\end{table*}

\subsection{Further Analysis}
\noindent\textbf{Effects of Backward Model and $z_e$.}
To gain an insight into the effectiveness of backward dialogue model and the latent variable $z_e$, we plot the \textit{Emotion accuracy} score of each emotion label based on Sing-Emp-Paired and Dual-Emp-Paired in Figure \ref{fig:ana1}. As we can see, for Sing-Emp-Paired, some emotion categories can achieve pretty high accuracy, but in general, the accuracy is unbalanced among all emotions, which indicates that $z_e$ cannot construct the emotion consensus well by only considering the contexts. In contrast, Dual-Emp-Paired not only improves the overall \textit{Emotion accuracy}, but also exhibits a relatively even performance over all 32 emotions. Therefore, $z_e$ can better understand the emotion via capturing emotion consensus with both forward and backward dialogue models.

\noindent\textbf{Choices of the Unpaired Data.}
The threshold $s$ we use in previous experiments equals to 0.60. Here, we choose different options to show their influence on the empathetic dialogue generation task. Table \ref{tab:s} shows that more unpaired emotional data does not lead to better results as some labels are not adequate with a low confidence. The emotion classifier we applied to label the utterances from Reddit and Twitter is based on a 32-category classification task, thus it is hard to get very accurate results. Though the predicted emotion labels are noisy, these samples are good enough to train our model in practice.

\section{Conclusion and Future Work}
In this paper, we propose a dual-generative model, Dual-Emp, to generate the empathetic response given a context. We point out that conducting an empathetic conversation is a bidirectional process, and empathy is mainly reflected by emotion consensus between the context and response. Then we couple forward and backward dialogue models with a discrete latent variable denoting the emotion consensus. Moreover, we integrate unpaired emotional data from open-domain conversations into Dual-Emp to relieve the need of paired data. Experimental results on a benchmark dataset show that Dual-Emp can generate fluent, related, informative, and empathetic responses. As the future work, we will prove the effectiveness of our method based on pre-trained models, and analyze how classification errors in unpaired data affect the generation.

\section*{Acknowledgements}
We would like to thank all the reviewers for their insightful and valuable comments and suggestions.

\bibliography{anthology,emnlp2021_new}
\bibliographystyle{acl_natbib}

\end{document}